\documentclass[runningheads]{llncs}
\usepackage{float} 
\usepackage{minted}
\usepackage{subcaption}
\usepackage{makecell}  
\usepackage{amsmath}
\usepackage{amsfonts}

\usepackage{multirow}

\usepackage{tabularx}
\usepackage{makecell}
\usepackage{rotating} 
\usepackage{booktabs} 
\usepackage{adjustbox} 
\usepackage{lscape}
\usepackage{caption} 
\usepackage{graphicx}  
\usepackage[table,xcdraw]{xcolor}  
\usepackage{longtable}
\usepackage[T1]{fontenc}
\usepackage{pgfplots}
\pgfplotsset{compat=1.18}
\usepackage{hyperref}
\usepackage[moderate]{savetrees}

\begin{document}
\title{Actor-Enriched Time Series Forecasting of Process Performance}

\author{
Aurélie Leribaux\inst{1} \and
Rafael Oyamada \inst{1} \and
Johannes De Smedt\inst{1} \and
Zahra Dasht Bozorgi\inst{2} \and
Artem~Polyvyanyy\inst{2} \and 
Jochen De Weerdt\inst{1}
}
\authorrunning{A. Leribaux et al.}

\institute{
Research Centre for Information Systems Engineering, KU Leuven, Belgium
\and
The University of Melbourne, Australia
}

\maketitle              
\begin{abstract}
Predictive Process Monitoring (PPM) is a key task in Process Mining that aims to predict future behavior, outcomes, or performance indicators. Accurate prediction of the latter is critical for proactive decision-making. Given that processes are often resource-driven, understanding and incorporating actor behavior in forecasting is crucial. Although existing research has incorporated aspects of actor behavior, its role as a time-varying signal in PPM remains limited. 
This study investigates whether incorporating actor behavior information, modeled as time series, can improve the predictive performance of throughput time (TT) forecasting models. Using real-life event logs, we construct multivariate time series that include TT alongside actor-centric features, i.e., actor involvement, the frequency of continuation, interruption, and handover behaviors, and the duration of these behaviors. We train and compare several models to study the benefits of adding actor behavior. The results show that actor-enriched models consistently outperform baseline models, which only include TT features, in terms of RMSE, MAE, and R\textsuperscript{2}. These findings demonstrate that modeling actor behavior over time and incorporating this information into forecasting models enhances performance indicator predictions. 
\end{abstract}

\section{Introduction}\label{sec:introduction}

PPM enables organizations to anticipate future behavior and performance of ongoing business processes using historical execution data~\cite{PPM}. While most research in PPM focuses on predicting the outcome or remaining time of individual cases~\cite{PPM,camargo2019,efren2021}, considerably less attention has been given to forecasting process-level performance indicators, such as average daily throughput time (TT). Yet, several studies identify TT as among the most critical KPIs used in process mining dashboards and operational decision-making across various domains \cite{vidgof2023impactprocesscomplexityprocess,signavio}. This makes forecasting TT highly valuable for operational planning, resource allocation, and capacity management.

In parallel, resource information captured in event logs has been shown to influence process outcomes~\cite{SENDEROVICH2019255,KIM2022113669} and model performance \cite{navarin2017lstmnetworksdataawareremaining,10.1007/978-3-030-46633-6_4}. However, most existing approaches encode resources in a static or categorical way \cite{KIM2022113669}, neglecting the dynamic nature of actor behavior. Recent work has begun to explore richer behavioral abstractions~\cite{10680657}, but these efforts have largely remained static and descriptive with limited application to predictive modeling. As such, the potential of actor behavior as a time-varying signal in performance forecasting remains underexplored.

This paper addresses this gap by investigating whether incorporating actor behavior, modeled as a multivariate time series, improves the forecasting of daily TT in business processes. We extract interpretable actor-centric features, such as the duration and frequency of continuation, interruption, and handover behaviors, from real-life event logs and align them with historical TT values. We then use these features to train a variety of forecasting models, ranging from autoregressive baselines to machine learning and deep learning architectures.

Our main contributions are:
\begin{itemize}
    \item We propose a novel actor-enriched forecasting framework for processes that models actor behavior as time series features aligned with TT.
    \item We validate this framework on three real-life BPIC event logs from distinct domains and scales.
    \item We show empirically that actor behavior features consistently improve TT forecasting performance across several model families, especially tree-based learners and recurrent nets.
\end{itemize}

The remainder of this paper is structured as follows. Section~\ref{sec:background} introduces the necessary terms, methods, and related work. Then, Section~\ref{sec:methodology} thoroughly describes the methodology, followed by the results in Section~\ref{sec:results} and a discussion in section~\ref{sec:discussion}.  Finally, Section~\ref{sec:conclusion} concludes this paper.

\section{Background and Related Work} \label{sec:background}

This section introduces event logs and how actor behavior can be represented as time-varying features. We then position our work within PPM, focusing on daily TT forecasting as influenced by dynamic resource behavior.

\subsection{Event Logs and Actor Behavior} \label{sec:behavioral-classification}

Event logs are the primary data source in process mining and capture the execution history of business processes. In this work, each event in a log typically records which activity was performed, in what case, when the activity occurred, and which resource (or actor) executed it. Formally, an event can be represented as a tuple \( e = (c,\, a,\, t,\, r) \), where \( c \) is the case identifier, \( a \) is the activity, \( t \) is the timestamp, and \( r \in \mathcal{R} \) is the resource that executed the event. A \emph{resource} can refer to a human actor, system agent, or any entity responsible for activity execution. A case trace \( \sigma_c \) is a sequence of such events ordered by timestamp for a given process instance.

Among these attributes, the resource dimension plays a particularly important role in characterizing process behavior. Rather than representing static information about who performed which activity, we focus on how work transitions between actors. Specifically, we extract \emph{actor features} by analyzing pairs of consecutive events within the same case. Each such pair, or \emph{transition}, is classified into one of four behavior types \cite{leribaux2025linkingactorbehaviorprocess}: \emph{continuation (C)}, \emph{interruption (I)}, \emph{handover to idle (HI)}, and \emph{handover to busy (HB)}. These types describe whether a resource continues their own work, is interrupted by other cases, or passes work to another actor who may or may not already be busy.

To capture how these behaviors evolve over time, we transform them into daily time series. Let \( \mathcal{D} = \{d_1, d_2, \ldots, d_n\} \) denote the ordered sequence of days on which at least one case started in the event log, and \( \mathcal{B} = \{\mathsf{C}, \mathsf{I}, \mathsf{HI}, \mathsf{HB}\} \) denote the set of actor behavior types. Each transition is labeled with a behavior type \( b \in \mathcal{B} \) and associated with the date of its first event. For each day \( d \in \mathcal{D} \) and behavior \( b \), we define:

\begin{enumerate}
  \item Daily behavior count: The number of transitions of type \( b \) whose first event occurred on day \( d \)
  \[
  \mathcal{F}_b(d) = \left| \left\{ p \in P \mid \text{behavior}(p) = b \land \text{date}(p) = d \right\} \right|
  \]
  where \( P \) is the set of all transitions, \( \text{behavior}(p) \) returns the behavior type of transition \( p \), and \( \text{date}(p) \) is the date of its first event.

  \item Daily behavior duration: The total time (in seconds) spent in transitions of type \( b \) whose first event occurred on day \( d \)
  \[
  \mathcal{T}_b(d) = \sum_{\substack{p \in P \\ \text{behavior}(p) = b \\ \text{date}(p) = d}} \delta_t(p)
  \]
  where \( \delta_t(p) \) is the time in seconds between the two events in transition \( p \).
\end{enumerate}

These metrics produce time series that describe daily variations in actor involvement. Let \( T = \{1, \dots, N\} \) be the sequence of days. For each behavior type \( b \in \mathcal{B} \) and metric \( m \in \{\mathcal{F}, \mathcal{T}\} \), we define a univariate time series \( \mathcal{X}_b^{(m)} = \{x_b^{(m)}(t)\}_{t \in T} \), where \( x_b^{(m)}(t) = m_b(d_t) \). That is, \( \mathcal{X}_b^{(\mathcal{F})} \) captures the daily counts of transitions of type \( b \), and \( \mathcal{X}_b^{(\mathcal{T})} \) captures their total durations.

\paragraph{Comparison to Prior Work.} Prior work has highlighted the value of resource information in PPM, particularly for improving predictions of remaining time or process outcomes. For example, \cite{navarin2017lstmnetworksdataawareremaining} demonstrate that incorporating resource labels into LSTM models can improve accuracy, but their representation is limited to static one-hot encoded identifiers. More broadly, most approaches treat resources performing a task as a simple categorical variable \cite{KIM2022113669}. To address this limitation, \cite{KIM2022113669} propose embedding resource IDs alongside handcrafted features such as the frequency of activity execution, thereby introducing a limited notion of experience. Similarly, \cite{10.1007/978-3-030-46633-6_4} reduce the complexity of resource space using clustering, grouping resources based on shared attributes to improve generalization. While these approaches do not solely rely on categorical variables, they still operate on static resource features, without modeling how resources interact or transition across tasks in a process. A complementary line of research by \cite{SENDEROVICH2019255} introduces inter-case resource competition as a feature, modeling how concurrent cases affect resource load at the system level. While this captures dynamic effects, it focuses on system-level pressure rather than the fine-grained behaviors of individual actors. Crucially, none of these approaches capture how resources behave within a case,i.e., how tasks are handed off, interrupted, or continued at the actor level across successive events. 

This behavioral dynamic is the focus of our work. It captures how individual actors behave across consecutive events, as shown by the resulting daily series \( \mathcal{X}_b^{(\mathcal{F})} \) and \( \mathcal{X}_b^{(\mathcal{T})} \). We align these with TT, also modeled as time series, and let them serve as explanatory variables in forecasting models. While \cite{10680657} use the behavior types for interpretability, our approach leverages them for predictive modeling, forecasting how behavior may impact process performance.



\subsection{PPM and TT forecasting} \label{sec:eventlogs}

PPM is a traditional field in process mining that studies predictive models that aim to predict the future of ongoing processes \cite{PPM}. PPM concerns three main prediction tasks, i.e., case outcome, the next events in a running case, and performance-related predictions. In performance-related predictions, predicting the remaining time for ongoing cases has become a core task, enabling real-time decision support and service level agreement management \cite{camargo2019}. Therefore, most research in PPM has focused on individual case-level predictions \cite{efren2021}. 

The importance of TT as a performance KPI is well-established in both academic and applied contexts. Several studies identify TT as among the most critical KPIs used in process mining dashboards and operational decision-making across various domains \cite{vidgof2023impactprocesscomplexityprocess,signavio}. In this context, TT serves as a proxy for overall process efficiency.
Similarly, some studies in industrial and manufacturing contexts have emphasized the importance of TT forecasting. For example, forecasting TT along different stages of an order fulfillment process to support operational planning has been proposed \cite{hiller2023throughput}. Furthermore, recent work developed data-driven models to predict throughput bottlenecks in production environments \cite{SUBRAMANIYAN2018533}. 

Our work builds on these foundations by forecasting the average TT of new cases expected to start on the following day, offering a forward-looking perspective on process performance that complements existing real-time monitoring approaches. In process mining, TT is typically defined at the case level as the duration between the timestamp of the first and last event in a case $c$: $
TT(c) = t(e_n) - t(e_1)$, where $e_1$ and $e_n$ denote the first and last events of case $c$, respectively.

To forecast process-level performance over time, we aggregate TT values on a daily basis by grouping cases according to their start date (i.e., the date of $t(e_1)$). For each day $d \in D$, where $D$ is the sequence of all days on which at least one case started in the event log, we define the daily average TT as:
\[
\mathcal{TT}(d) = \frac{1}{|C_d|} \sum_{c \in C_d} TT(c),
\]
where $C_d$ is the set of completed cases from the event log that have a start timestamp (i.e., the timestamp of the first event in the case) on day $d$, and for which both a start and end timestamp are present (i.e., completed cases only). $TT(c)$ is defined as the time difference, in hours, between its first and last event. This results in a univariate time series $\{\mathcal{TT}(d)\}_{d \in D}$, where each value reflects the average case duration on a given day. We restrict our analysis to completed cases only, ensuring that both the start and end timestamps are available. This exclusion of ongoing cases prevents biases from incomplete traces. Similar to our earlier work, where $\mathcal{TT}$ served as target to test for causality with behavioral predictors \cite{leribaux2025linkingactorbehaviorprocess}, we treat $\mathcal{TT}$ as the primary variable to be forecasted, serving as a proxy to process performance. 


\section{Methodology}\label{sec:methodology}

This section describes the methodology used to investigate whether time-varying actor behavior improves the forecasting of process performance indicators, specifically $\mathcal{TT}$. 

\begin{figure}[htpb]
    \centering
  \includegraphics[width=\textwidth]{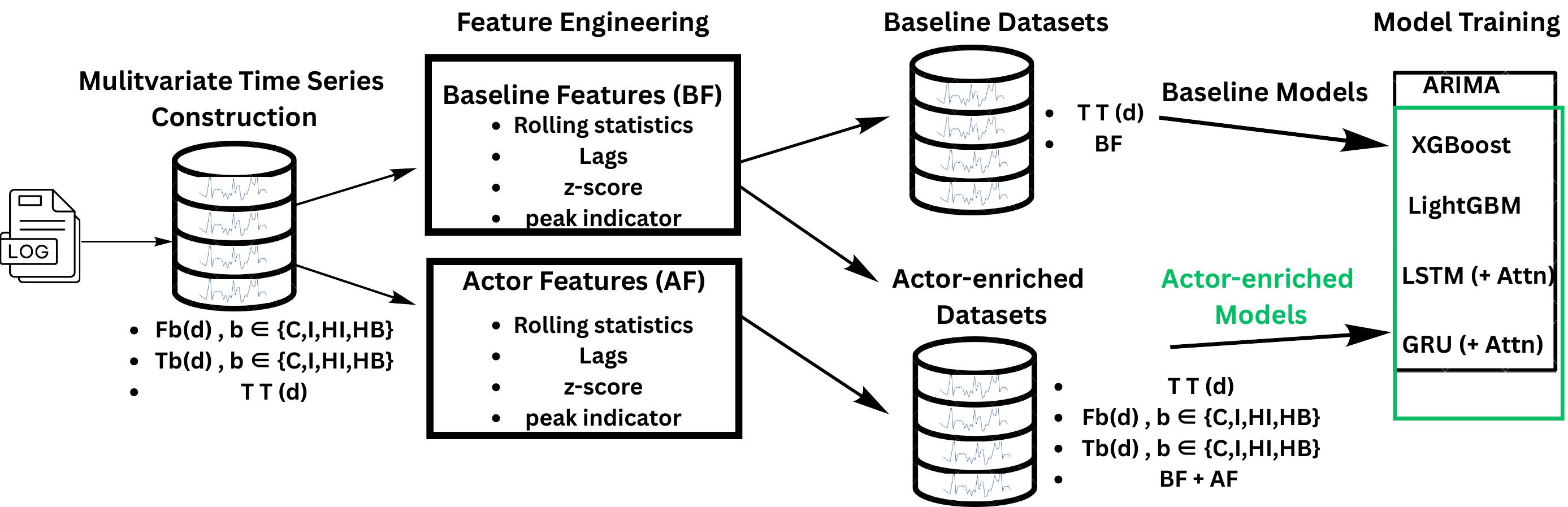}
    \caption{Overview of the methodology for incorporating actor behavior into TT prediction.}
    \label{fig:workflow}
\end{figure}

\paragraph{Multivariate Time Series Construction and Feature Engineering}\label{sec:featureengineering}

An overview of our approach is illustrated in Figure~\ref{fig:workflow}. We started from event logs from real-life business processes, where each event contains at least the case identifier, timestamp, activity label, and involved resource. Then, enanhanced event logs were used, which complement the events with the type of actor behavior, and the duration of that behavior. Subsequently, the univariate time series $\{\mathcal{TT}(d)\}_{d \in D}$, \( \mathcal{F}_b(d) \), and \( \mathcal{T}_b(d) \), introduced in the previous section, are extracted from these event logs. This way, we construct a multivariate time series dataset at a daily time step, where each time step refers to the aggregated (via averaging) behaviors. 

We aim to forecast the average TT of cases scheduled to start on the subsequent day. To accomplish this, we construct two categories of time series features. The first features, referred to as \emph{baseline features}, are derived solely from historical TT values. These include daily lagged values (1 to 20 days), rolling statistics (mean, standard deviation, and maximum) over windows of 3, 7, and 14 days, a 7-day z-score, and a peak indicator. The peak indicator is set to one on days where the TT series shows a local maximum, detected using a prominence-based algorithm with a minimum distance of 7 days between peaks \cite{scipy_find_peaks}. The second features category, the \emph{actor-enriched features}, encapsulates the time-varying actor behaviors \( \mathcal{F}_b(d) \) and \( \mathcal{T}_b(d) \). The actor-enriched features model both the frequency and duration of interactions between resources, i.e., the number and duration of continuation actions, interruptions, and handovers, distinguished by whether they occur to idle or busy resources. They are engineered similarly to the baseline features, computed across all cases per day (i.e., aggregated inter-case). Based on the engineered features, we generate two multivariate time series datasets: the baseline dataset, containing only the baseline features, and the actor-enriched dataset, which augments the baseline with actor features.

\begin{figure}[htpb]
    \centering
    \includegraphics[width=\textwidth, trim=0cm 6cm 0cm 6cm]{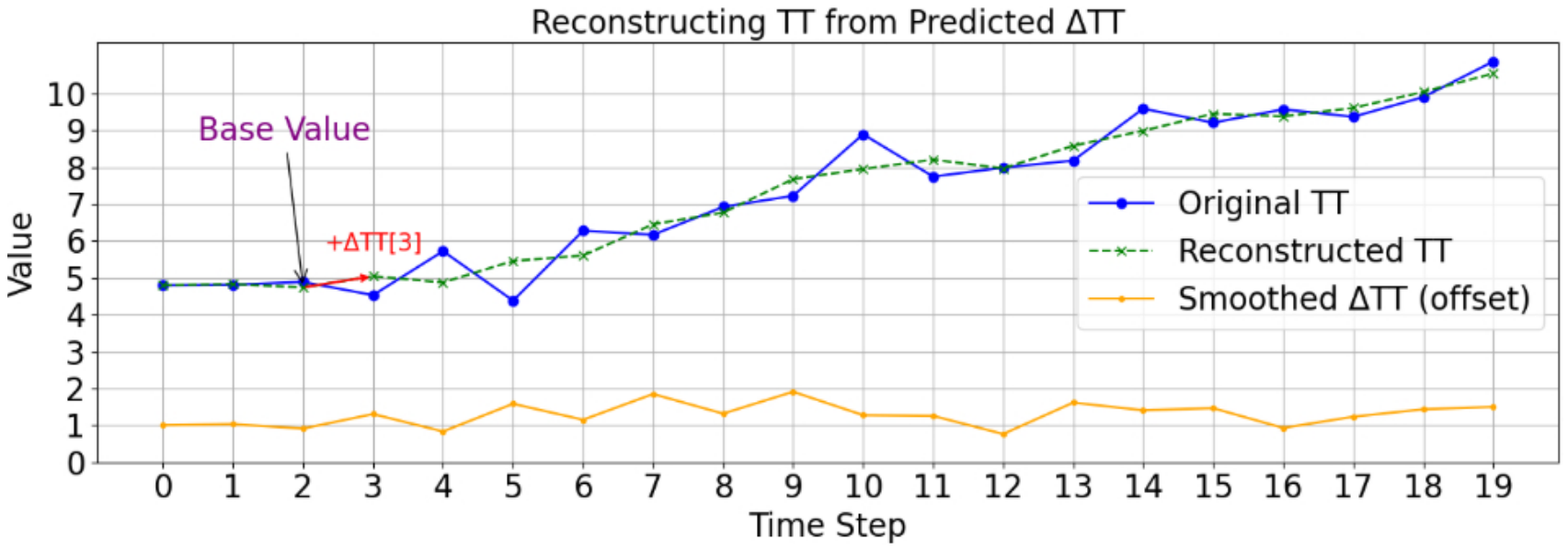}
    \caption{Illustration of reconstructing the target variable TT by incrementally adding predicted smoothed $\Delta TT$ values to a base value.}
    \label{fig:reconstruct_tt}
\end{figure}

To improve temporal learning stability, we predict the \emph{daily, smoothed first difference of TT} ($\Delta TT$), computed via a 3-point rolling average. Final predictions are reconstructed by adding the predicted $\Delta TT$ to the previous day's TT (the base value).  Figure~\ref{fig:reconstruct_tt} illustrates this process, where the red arrow, for example, shows how $\Delta TT[3]$ updates the TT at step 2. This approach is inspired by residual learning strategies such as R2N2~\cite{goel2017r2n2residualrecurrentneural}, which first model a time series with a simple linear method and then predict the residuals using a neural network. Similarly, predicting $\Delta TT$ simplifies the task by focusing the model on short-term variation rather than the full TT trajectory.

\paragraph{Model Training}

We compare three model classes with distinct characteristics using both baseline and actor-enriched feature sets. We aim to assess if it is possible to improve predictive performances via our enriched datasets, regardless of the employed algorithm. Thus, as a benchmark, an ARIMA model is solely trained on historical first differences of TT without incorporating any actor behavior or contextual features, reflecting a random walk. Then, Gradient Boosted Trees (XGBoost, LightGBM) are trained on structured features to forecast $\Delta$TT. Finally, we design hybrid deep learning models combining Conv1D layers for local pattern extraction, bidirectional RNNs (GRU or LSTM), and optional attention. Convolutional layers help to capture short-term temporal dynamics, complementing the RNNs' ability to model both long-term and short-term dependencies~\cite{bai2018empirical}. All models are trained with early stopping and learning rate scheduling. Predictions are made on standardized $\Delta$TT and reconstructed to full TT through an inverse transformation.

\section{Experimental Evaluation}\label{sec:results}

We evaluate our approach using three real-life event logs from the Business Process Intelligence Challenge (BPIC) series: BPIC2017\footnote{\url{https://data.4tu.nl/articles/dataset/BPI_Challenge_2017/12696884}}, BPIC2012\footnote{\url{https://data.4tu.nl/articles/dataset/BPI_Challenge_2012/12689204}}, and BPIC2011 (Hospital log)\footnote{\url{https://data.4tu.nl/articles/dataset/Real-life_event_logs_-_Hospital_log/12716513}}. These datasets span two diverse domains including financial services, and healthcare. They vary in TT granularity (hours vs. days), data volume, and behavioral complexity, making them well-suited for evaluating model robustness under different real-world conditions.

Our experimental design is guided by the hypothesis that actor behavior, particularly how resources handle task handovers, interruptions, and continuations, influences process performance over time as measured by daily TT. This hypothesis is motivated by prior work showing that TT is one of the key performance indicators in business process management \cite{vidgof2023impactprocesscomplexityprocess}, and that the way resources manage work transitions directly affects execution delays and process efficiency ~\cite{10680657}. 
This setup allows us to address the following research question: \textit{to what extent do resource behavior profiles improve the forecasts of daily TT in business processes?}


\subsection{Setup}\label{sec:setup}
To address the research question, we construct for each dataset a multivariate time series that includes the daily average TT as the target variable, along with several actor-related features engineered from resource transitions. We implemented the approach in Python, and the code is available on our GitHub repository\footnote{https://github.com/aurelieleribaux-1/actor-behavior-TT-forecasting/tree/main}. The following sections explain the setup and the results.

Each dataset is then split chronologically into an 80\% training set and a 20\% holdout test set for final evaluation. We apply five-fold time series cross-validation, ensuring that training data always precedes test data in each fold to preserve temporal order and prevent data leakage.

Hyperparameters for all models (i.e., XGBoost, LightGBM, LSTM/GRU with and without attention) are selected from a predefined grid via time-series cross-validation. For the tree-based models, we tune the number of estimators (\{1000, 1500, 3000\}), learning rate (\{0.05, 0.1, 0.2\}), maximum tree depth (\{5, 6, 7\}), feature fraction (\{0.6, 0.8, 0.9\}), and bagging fraction (\{0.6, 0.8, 0.9, 1.0\}). For the neural models, we jointly tune RNN and CNN components: hidden units (\{64, 128, 256, 512\}), dense units (\{32, 64, 128, 256\}), dropout (\{0.2, 0.3\}), batch size (\{16, 32\}), and CNN kernel/filter/pool configurations (\{3, 5\}, \{32, 64\}, \{2, 3\}). All models are trained to minimize RMSE on validation folds, and the best configuration is retrained on the full training set before evaluation. Rather than performing separate optimization for each model variant (baseline vs.\ actor-enriched), we used a practical strategy: we explored a shared grid of plausible configurations and retained those under which actor-enriched models consistently outperformed their baselines. This approach does not guarantee globally optimal tuning for each variant but enables a fair comparison under matched conditions, while keeping computational cost tractable. It also ensures that observed gains can be attributed to actor-related features rather than differences in model capacity or training dynamics.

To isolate the added value of actor-related features, we evaluate and compare holdout test results of the baseline and actor-enriched models using the RMSE, MAE, and (R\textsuperscript{2}) metrics. Furthermore, an ARIMA model is trained as a baseline model. Moreover, to better understand which input features most influence the forecasts, we used both model-specific and model-agnostic feature importance techniques. SHAP values were used for tree-based models to capture the marginal contribution of each feature. For RNNs, we applied permutation importance based on RMSE, which quantifies the increase in prediction error when a feature's values are randomly shuffled.


\begin{table}[htbp]
\centering
\caption{Selected hyperparameter configurations per model and dataset, selected via grid search on time-series cross-validation.}
\label{tab:final_hyperparams}
\resizebox{\textwidth}{!}{%
\begin{tabular}{l
ccc  
ccc  
ccc  
ccc  
ccc  
ccc  
ccc  
ccc  
ccc  
ccc  
}
\toprule
\multicolumn{1}{c}{} 
& \multicolumn{3}{c}{\textbf{n\_estimators}} 
& \multicolumn{3}{c}{\textbf{Learning Rate}} 
& \multicolumn{3}{c}{\textbf{Max Depth}} 
& \multicolumn{3}{c}{\textbf{Feature Fraction}} 
& \multicolumn{3}{c}{\textbf{Bagging Fraction}} \\
\cmidrule(lr){2-4} \cmidrule(lr){5-7} \cmidrule(lr){8-10} \cmidrule(lr){11-13} \cmidrule(lr){14-16}
\textbf{Model} & BPI17 & BPI12 & BPI11 & BPI17 & BPI12 & BPI11 & BPI17 & BPI12 & BPI11 & BPI17 & BPI12 & BPI11 & BPI11 & BPI12 & BPI11 \\
\midrule
XGBoost   & 1000 & 1000 & 3000 & 0.1 & 0.1 & 0.1 & 5 & 5 & 6 & 0.9 & 0.9 & 0.9 & \makebox[2cm][c]{0.9} & \makebox[2cm][c]{0.9} & \makebox[2cm][c]{0.9} \\
LightGBM  & 1500 & 1500 & 1500 & 0.05 & 0.05 & 0.2 & 5 & 5 & 7 & 0.9 & 0.9 & 0.6 & \makebox[2cm][c]{0.9} & \makebox[2cm][c]{0.9} & \makebox[2cm][c]{0.8} \\
\midrule
\midrule
\multicolumn{1}{c}{} 
& \multicolumn{3}{c}{\textbf{Hidden Units}} 
& \multicolumn{3}{c}{\textbf{Dropout}} 
& \multicolumn{3}{c}{\textbf{Dense Units}} 
& \multicolumn{3}{c}{\textbf{Batch Size}} 
& \multicolumn{3}{c}{\textbf{CNN (Kernel–Filters–Pool)}} \\
\cmidrule(lr){2-4} \cmidrule(lr){5-7} \cmidrule(lr){8-10} \cmidrule(lr){11-13} \cmidrule(lr){14-16}
\textbf{Model} & 2017 & 2012 & 2011 & 2017 & 2012 & 2011 & 2017 & 2012 & 2011 & 2017 & 2012 & 2011 & 2017 & 2012 & 2011 \\
\midrule
LSTM         & 64 & 64 & 128 & 0.2 & 0.3 & 0.2 & 32 & 32 & 32 & 16 & 32 & 32 & \makebox[2cm][c]{3-64-2} & \makebox[2cm][c]{3-64-2} & \makebox[2cm][c]{3-64-2} \\
GRU          & 32 & 64 & 128 & 0.2 & 0.3 & 0.2 & 32 & 64 & 64 & 16 & 32 & 32 & \makebox[2cm][c]{3-64-2} & \makebox[2cm][c]{3-64-2} & \makebox[2cm][c]{3-64-2} \\
LSTM (Attn)  & 64 & 64 & 512 & 0.2 & 0.2 & 0.3 & 32 & 32 & 128 & 32 & 16 & 32 & \makebox[2cm][c]{3-64-2} & \makebox[2cm][c]{3-64-2} & \makebox[2cm][c]{3-64-2} \\
GRU (Attn)   & 128 & 64 & 256 & 0.2 & 0.2 & 0.3 & 32 & 32 & 128 & 32 & 16 & 32 & \makebox[2cm][c]{3-64-2} & \makebox[2cm][c]{3-64-2} & \makebox[2cm][c]{3-64-2} \\
\bottomrule
\end{tabular}%
}
\end{table}

\begin{table}[htbp]
\centering
\small
\caption{Final Holdout Test Results on BPIC 2017, 2012, and 2011 Datasets with 95\% Bootstrapped Confidence Intervals. For BPIC 2017 and BPIC 2012, errors are reported in hours. For BPIC 2011, errors are reported in days.}
\label{tab:final_results_all}
\resizebox{\textwidth}{!}{%
\begin{tabular}{l|c|c|c|c|c|c|c|c|c}
\toprule
\textbf{Model} 
& \multicolumn{2}{c|}{\textbf{RMSE}} &  
& \multicolumn{2}{c|}{\textbf{MAE}} &  
& \multicolumn{2}{c|}{\textbf{R\textsuperscript{2}}} & \\
& Baseline & Actor & \textbf{$\Delta$ }  
& Baseline & Actor & \textbf{$\Delta$ }
& Baseline & Actor & \textbf{$\Delta$}  \\
\midrule
\multicolumn{10}{c}{\textbf{BPIC 2017}} \\
\midrule
ARIMA (Benchmark) & 14.5125 & -- & -- & 10.8879 & -- & -- & -- & -- & -- \\
XGBoost        & 10.052 ± 1.327 & 9.162 ± 0.931 & 0.889 & 7.818 ± 0.950 & 7.621 ± 0.759 & 0.197 & 0.943 ± 0.024 & 0.953 ± 0.016 & 0.011 \\
LightGBM       & 11.284 ± 1.376 & 9.226 ± 0.923 & 2.058 & 8.775 ± 1.081 & 7.477 ± 0.790 & 1.298 & 0.928 ± 0.030 & 0.953 ± 0.015 & 0.025 \\
LSTM (w/ Attn) & 13.391 ± 1.813 & 12.689 ± 1.590 & 0.701 & 10.523 ± 1.543 & 10.161 ± 1.410 & 0.362 & 0.866 ± 0.045 & 0.880 ± 0.040 & 0.013 \\
GRU (w/ Attn)  & 13.345 ± 1.791 & 13.080 ± 1.642 & 0.265 & 10.469 ± 1.538 & 10.376 ± 1.448 & 0.093 & 0.868 ± 0.044 & 0.869 ± 0.053 & 0.002 \\
LSTM           & 13.075 ± 1.732 & 12.640 ± 1.489 & 0.436 & 10.254 ± 1.513 & 9.989 ± 1.422 & 0.264 & 0.873 ± 0.042 & 0.881 ± 0.040 & 0.008 \\
GRU            & 12.891 ± 1.482 & 12.798 ± 1.601 & 0.093 & 10.110 ± 1.492 & 10.079 ± 1.470 & 0.031 & 0.876 ± 0.040 & 0.878 ± 0.040 & 0.002 \\
\midrule
\multicolumn{10}{c}{\textbf{BPIC 2012}} \\
\midrule
ARIMA (Benchmark) & 15.8513 & -- & -- & 12.03839 & -- & -- & -- & -- & -- \\
XGBoost        & 12.548 ± 2.490 & 9.616 ± 2.338 & 2.932 & 9.031 ± 2.063 & 7.256 ± 1.637 & 1.775 & 0.853 ± 0.146 & 0.911 ± 0.093 & 0.058 \\
LightGBM       & 13.052 ± 2.341 & 12.810 ± 2.787 & 0.242 & 10.848 ± 1.777 & 9.913 ± 2.092 & 0.935 & 0.841 ± 0.156 & 0.855 ± 0.136 & 0.014 \\
LSTM (w/ Attn) & 17.813 ± 2.687 & 13.615 ± 2.957 & 4.198 & 13.981 ± 2.610 & 10.381 ± 2.099 & 3.600 & 0.789 ± 0.122 & 0.869 ± 0.100 & 0.080 \\
GRU (w/ Attn)  & 17.521 ± 2.695 & 12.825 ± 3.032 & 4.696 & 13.568 ± 2.629 & 9.750 ± 2.051 & 3.819 & 0.796 ± 0.120 & 0.884 ± 0.093 & 0.088 \\
LSTM           & 17.751 ± 2.742 & 13.354 ± 2.836 & 4.396 & 13.897 ± 2.625 & 9.886 ± 2.149 & 4.011 & 0.791 ± 0.122 & 0.875 ± 0.093 & 0.085 \\
GRU            & 17.736 ± 2.733 & 13.910 ± 2.919 & 3.827 & 13.932 ± 2.612 & 10.315 ± 2.233 & 3.617 & 0.791 ± 0.122 & 0.863 ± 0.105 & 0.072 \\
\midrule
\multicolumn{10}{c}{\textbf{BPIC 2011}} \\
\midrule
ARIMA (Benchmark) & 48.3250 & -- & -- & 43.4461 & -- & -- & -- & -- & -- \\
XGBoost        & 22.457 ± 1.317 & 21.204 ± 1.048 & 1.253 & 17.275 ± 1.021 & 15.993 ± 0.922 & 1.283 & 0.882 ± 0.020 & 0.895 ± 0.017 & 0.013 \\
LightGBM       & 20.562 ± 1.233 & 20.017 ± 1.059 & 0.545 & 15.676 ± 0.908 & 15.253 ± 0.891 & 0.423 & 0.901 ± 0.018 & 0.907 ± 0.016 & 0.005 \\
LSTM (w/ Attn) & 23.017 ± 1.396 & 22.425 ± 1.398 & 0.592 & 16.537 ± 1.144 & 15.205 ± 1.176 & 1.332 & 0.838 ± 0.021 & 0.846 ± 0.021 & 0.008 \\
GRU (w/ Attn)  & 24.112 ± 1.399 & 22.450 ± 1.305 & 1.662 & 18.212 ± 1.120 & 17.198 ± 1.023 & 1.014 & 0.821 ± 0.026 & 0.845 ± 0.020 & 0.024 \\
LSTM           & 18.021 ± 0.780 & 22.425 ± 1.398 & -4.405 & 14.865 ± 0.694 & 15.205 ± 1.176 & -0.340 & 0.900 ± 0.014 & 0.846 ± 0.021 & -0.054 \\
GRU            & 24.112 ± 1.399 & 22.450 ± 1.305 & 1.662 & 18.212 ± 1.120 & 17.198 ± 1.023 & 1.014 & 0.821 ± 0.026 & 0.845 ± 0.020 & 0.024 \\
\bottomrule
\end{tabular}%
}
\end{table}

\subsection{Results}

The selected hyperparameters (Table~\ref{tab:final_hyperparams}) diverge clearly from defaults. For instance, XGBoost defaults to a learning rate of 0.3 and 100 estimators, whereas our best configurations use 0.05 to 0.1 with 1000 to 3000 estimators. LightGBM similarly defaults to 0.1 and 100 estimators, while our tuned models use up to 1500. For RNNs, common defaults suggest 128 or 256 hidden units with no or high dropout (e.g., 0.5), yet our best results use 32 to 128 units and dropout between 0.2 and 0.3. These differences led to significant validation gains. In particular, actor-enriched models under default settings often showed no advantage over baselines, indicating that tuning was essential for revealing their contribution. Despite the added computational cost, tuning was necessary for fair and informative comparisons. The results are shown in Table~\ref{tab:final_results_all}. Importantly, the models are trained to predict $\Delta TT$, then the final predicted TT values are reconstructed by adding the $\Delta TT$ predictions to the base values. Hence, all reported errors in the tables are computed on the latter reconstructed final TT predictions. This ensures that all metrics reflect performance on the original target variable in its natural unit. 

Across all datasets and model types, incorporating actor behavior features consistently improves predictive performance compared to baseline models. Actor-enriched pipelines achieve lower RMSE and MAE and higher R\textsuperscript{2} in nearly all cases. These improvements are most pronounced in BPIC 2012. In BPIC 2011, improvements are present but less consistent, likely due to the dataset’s larger size, higher variance, and more complex case structure. In all settings, actor-enriched models significantly outperform the ARIMA benchmark. This highlights the added value of structured, resource-aware features in TT forecasting.

Among the evaluated methods, tree-based models, especially XGBoost, consistently achieve the best overall performance. In BPIC 2012, for example, actor-enriched XGBoost reduces RMSE by nearly 2.93 hours, MAE by 1.78 hours, and improves R\textsuperscript{2} by 5.8 percentage points compared to its baseline counterpart. These models show more robust and statistically significant improvements, with narrower confidence intervals and larger margins over their baselines. This suggests that tree-based approaches are particularly effective at capturing both temporal and resource-driven patterns, due to their capacity to handle complex interactions and structured feature inputs. 
Accordingly, recurrent models such as LSTM and GRU also benefit from actor enrichment, though the improvements tend to be smaller and more sensitive to dataset characteristics. In BPIC 2012, attention-enhanced LSTM achieves an RMSE reduction of over 4.4 hours and an R\textsuperscript{2} improvement of 8.5 percentage points. However, in BPIC 2011, some RNN variants, especially plain LSTM, show less consistent performance. Confidence intervals often overlap, suggesting that while these models can learn from actor features, they may be less stable in more complex, high-variance environments. 
Therefore, this analysis successfully answers our research question by showcasing the effectiveness of incorporating actor-centric information for TT prediction regardless of the employed model.

\begin{figure}[htbp]
  \centering
  \begin{subfigure}[b]{0.48\textwidth}
    \includegraphics[width=\textwidth]{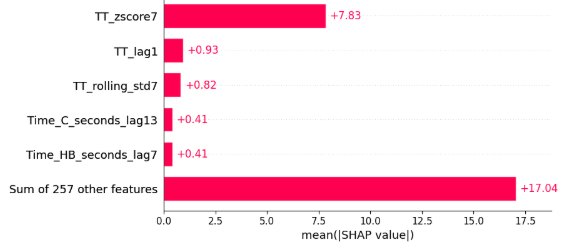}
    \caption{XGBoost – BPIC 2017}
  \end{subfigure}
  \hfill
  \begin{subfigure}[b]{0.48\textwidth}
    \includegraphics[width=\textwidth]{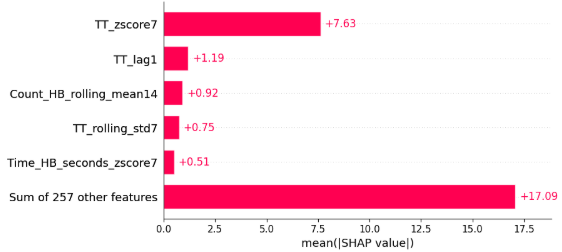}
    \caption{LightGBM – BPIC 2017}
  \end{subfigure}

  \begin{subfigure}[b]{0.48\textwidth}
    \includegraphics[width=\textwidth]{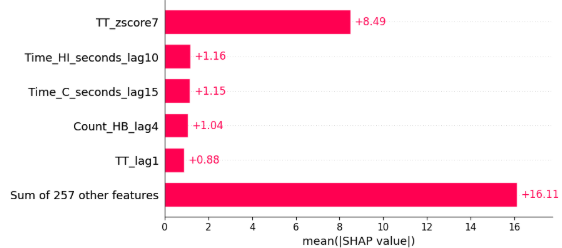}
    \caption{XGBoost – BPIC 2012}
  \end{subfigure}
  \hfill
  \begin{subfigure}[b]{0.48\textwidth}
    \includegraphics[width=\textwidth]{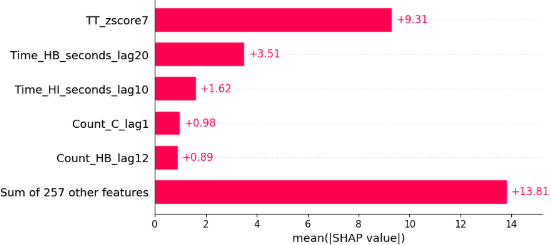}
    \caption{LightGBM – BPIC 2012}
  \end{subfigure}

  \begin{subfigure}[b]{0.48\textwidth}
    \includegraphics[width=\textwidth]{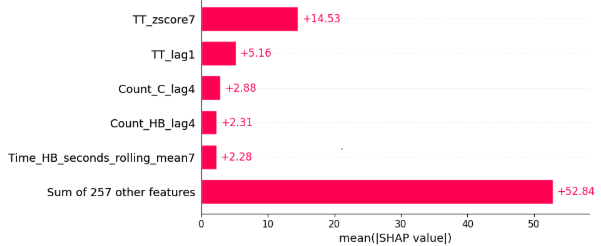}
    \caption{XGBoost – BPIC 2011}
  \end{subfigure}
  \hfill
  \begin{subfigure}[b]{0.48\textwidth}
    \includegraphics[width=\textwidth]{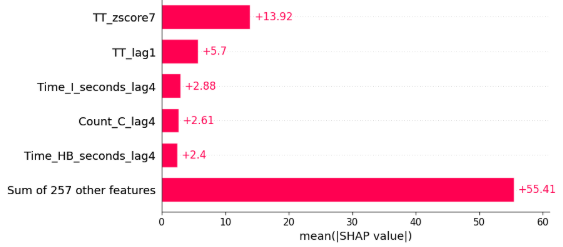}
    \caption{LightGBM – BPIC 2011}
  \end{subfigure}

  \caption{Top 10 most important predictive features (by mean SHAP value) for actor-enriched XGBoost and LightGBM models across all datasets.}
  \label{fig:shap_summary}
\end{figure}

Additionally, these features naturally lead to another benefit regarding the model's interpretability. Concerning the tree-based methods, Figure~\ref{fig:shap_summary} shows the top 5 most important predictive features as measured by mean SHAP values across all test samples for both XGBoost and LightGBM actor-enriched models for all datasets. Across all cases, the most influential features are derived from the target time series itself, particularly lagged values and z-score transformations such as \texttt{TT\_zscore7} and \texttt{TT\_lag1}, demonstrating strong temporal dependence. However, actor-related variables (e.g., \texttt{Count\_C\_lag4}, \texttt{Time\_HB\_seconds\_lag4}) also rank highly, confirming their added value in capturing workload and behavioral patterns that influence TT. Notably, the most predictive features remain relatively consistent across datasets and model types.

Regarding our deep neural nets, we employ the permutation importance technique to interpret the impact of features on each model.
Table~\ref{tab:perm_importance_all} shows the top 5 most important features of the predictions for the actor-enriched models across all datasets, measured as average change in RMSE. Interestingly, the results show a strong reliance on the actor features. Lagged and rolling indicators of handovers (\texttt{Count\_HB\_lag13}, \texttt{Count\_HB\_seconds\_lag8}) and continuations (\texttt{Count\_C\_rolling\_mean3}) frequently appear among the most important predictors. Moreover, several high-ranking features involve time-based behavior metrics (e.g., \texttt{Time\_HI\_seconds\_rolling\_max7}), suggesting that RNNs are particularly sensitive to time-varying fluctuations in workload and coordination activity. These findings reinforce the relevance of capturing dynamic actor behavior and show that RNNs can internalize such information when provided in well-engineered temporal formats. Notably, the top features vary slightly more across datasets compared to the SHAP-based rankings, indicating that RNNs may adapt more flexibly to dataset-specific dynamics.

\begin{table}[htbp]
\centering
\footnotesize
\caption{Top 5 most important features for actor-enriched RNN models across datasets, ranked by increase in RMSE when shuffled (permutation importance).}
\label{tab:perm_importance_all}
\begin{tabularx}{\textwidth}{l X r X r}
\toprule
\textbf{Dataset} & \multicolumn{2}{c}{\textbf{LSTM}} & \multicolumn{2}{c}{\textbf{GRU}} \\
\cmidrule(lr){2-3} \cmidrule(lr){4-5}
 & Feature & $\Delta$RMSE & Feature & $\Delta$RMSE \\
\midrule
\multirow{5}{*}{BPIC 2017}
& Time\_I\_seconds\_rolling\_mean7  & 0.366 & Time\_I\_seconds\_lag12         & 0.078 \\
& Time\_I\_seconds\_rolling\_max7   & 0.365 & Time\_I\_seconds\_rolling\_std7 & 0.067 \\
& Time\_I\_seconds\_rolling\_std7   & 0.328 & Time\_I\_seconds\_lag15         & 0.053 \\
& Time\_I\_seconds\_rolling\_mean14 & 0.312 & Time\_I\_seconds\_lag8          & 0.053 \\
& Time\_I\_seconds\_lag3            & 0.226 & Time\_I\_seconds\_lag11         & 0.051 \\
\midrule
\multirow{5}{*}{BPIC 2012}
& Time\_I\_seconds\_rolling\_max7 & 0.298 & Time\_HB\_seconds\_lag8       & 0.161 \\
& Time\_I\_seconds\_lag9         & 0.241 & Time\_C\_seconds\_lag17       & 0.160 \\
& Count\_C\_lag6                 & 0.216 & TT\_lag1                      & 0.149 \\
& Time\_HI\_seconds\_lag19       & 0.216 & Count\_HB\_lag13              & 0.139 \\
& Time\_HI\_seconds\_lag9        & 0.202 & Count\_C\_lag18               & 0.136 \\
\midrule
\multirow{5}{*}{BPIC 2011}
& Count\_HB\_lag8            & 0.024 & Count\_C\_rolling\_mean3         & 0.246 \\
& Count\_HI\_rolling\_mean3  & 0.020 & Count\_HI\_rolling\_max14        & 0.202 \\
& Time\_HB\_seconds\_lag7    & 0.019 & Time\_HB\_seconds\_rolling\_max14 & 0.201 \\
& Count\_HI\_lag19           & 0.017 & Count\_C\_lag2                  & 0.198 \\
& Count\_C                   & 0.016 & Count\_HB\_lag1                 & 0.167 \\
\bottomrule
\end{tabularx}
\end{table}

\section{Discussion} \label{sec:discussion}

This study demonstrates that incorporating actor behavior features into time-series models consistently improves TT forecasting across diverse real-world processes. The empirical evaluation on three BPIC datasets, spanning administrative (BPIC2017, BPIC2012) and clinical (BPIC2011) domains, shows that actor enrichment yields positive performance gains across multiple model families and error metrics, thereby confirming our research question. The use of three datasets with varying characteristics supports the generalizability of the findings: they span two different domains, data scale, TT distribution (hours vs.\ days), and process complexity, providing a robust testbed for evaluating predictive modeling under different conditions.

Across all datasets, tree-based models, i.e., XGBoost and LightGBM, consistently deliver the strongest results. They not only achieve the lowest error rates but also demonstrate more robust improvements when enriched with actor behavior features. This can be attributed to their ability to handle structured, aggregated features effectively and to model complex interactions through gradient boosting. Moreover, they offer additional advantages, such as lower training time, being less sensitive to hyperparameter tuning, generalizing well with limited assumptions about input structure. These properties make them particularly suitable for domains where engineered features, such as resource counts, durations, or rolling statistics, encode meaningful temporal and contextual signals. However, the relative gains for RNNs are more variable, especially for BPIC2011. This dataset presents longer, noisier, and more irregular process traces, which may challenge sequence models that rely on stable temporal dependencies. Additionally, the actor features used are aggregated at a daily level, which may reduce their alignment with step-wise recurrence. In such contexts, RNNs may not always extract full value from these signals. Notably, in BPIC2011, the LSTM model without attention performs worse with actor enrichment. Nonetheless, GRU variants generally remain robust, with GRU (with attention) showing consistent improvements across all datasets.

Despite these promising results, several limitations must be acknowledged. First, the prediction target is defined as a smoothed, differenced version of TT ($\Delta$TT), which may limit interpretability and sensitivity to extreme case durations. Second, all models operate on a fixed daily temporal resolution, which may overlook long-term fluctuations relevant in high-frequency process environments. Third, actor behavior features are hand-engineered using predefined templates (e.g., counts, durations, rolling statistics), potentially missing latent or nonlinear interaction patterns that could be learned automatically.  Finally, we did not fully optimize each model variant independently. Instead, we retained shared configurations where actor-enriched models consistently outperformed baselines. This pragmatic choice allowed us to assess feature contributions under comparable conditions, but may have limited baseline performance and confounded effect attribution.

Future work could explore richer actor representations, such as learned graph-based structures, and integrate multiple temporal resolutions. Additionally, models could be further trained and investigated to incorporate actor features into several KPI forecasting scenarios. Moreover, hybrid models that, for example, combine the structure-awareness of trees with the sequence modeling capacity of RNNs may offer a promising direction. Lastly, future work could tune each variant separately or fix one model (e.g., XGBoost) to more precisely evaluate the impact of actor features.

\section{Conclusion}  \label{sec:conclusion}

This paper investigates whether time-varying actor behavior can improve process-level performance forecasting, specifically the daily average TT. 
We constructed multivariate time series from real-life event logs that included actor behavior features such as the frequency and duration of continuation, interruption, and handover behavior, alongside historical TT. We trained and evaluated a range of forecasting models, i.e., ARIMA, gradient boosted trees (XGBoost, LightGBM), and recurrent neural networks (LSTM, GRU), on three BPIC datasets from different domains.
The results show that actor-enriched models consistently outperform their baseline counterparts across all datasets and model families. Tree-based models, particularly LightGBM and XGBoost, achieved the most stable and significant improvements, while RNN-based models also benefited strongly from actor information. These findings confirm that modeling actor behavior over time contributes positively to process performance forecasting and provides a new perspective on resource-aware predictive monitoring.

\bibliographystyle{splncs04}
\bibliography{bibliography}   
\end{document}